\documentclass[sigconf]{acmart}

\usepackage{booktabs} 

\setcopyright{none}

\acmDOI{}

\acmISBN{}

\acmConference[Explainable Robotic Systems Workshop, ACM Human-Robot Interaction conference]{Explainable Robotic Systems Workshop, ACM Human-Robot Interaction conference}{March 2018}{Chicago, IL USA}
\acmYear{2018}

\begin{document}

\title{Explain Yourself: A Natural Language Interface for Scrutable Autonomous Robots}

\author{Francisco J. Chiyah Garcia}
\affiliation{%
 \institution{Heriot-Watt University}
  \city{Edinburgh} 
  \country{UK} 
  \postcode{EH14 4AS}
}
\email{fjc3@hw.ac.uk}

\author{David A. Robb}
\affiliation{%
 \institution{Heriot-Watt University}
  \city{Edinburgh} 
  \country{UK} 
  \postcode{EH14 4AS}
}
\email{d.a.robb@hw.ac.uk}

\author{Xingkun Liu}
\affiliation{%
 \institution{Heriot-Watt University}
  \city{Edinburgh} 
  \country{UK} 
  \postcode{EH14 4AS}
}
\email{x.liu@hw.ac.uk}

\author{Atanas Laskov}
\affiliation{%
  \institution{SeeByte Ltd}
  \city{Edinburgh} 
  \country{UK} 
  \postcode{EH4 2HS}
}
\email{atanas.laskov@seebyte.com}

\author{Pedro Patron}
\affiliation{%
  \institution{SeeByte Ltd}
  \city{Edinburgh} 
  \country{UK} 
  \postcode{EH4 2HS}
}
\email{pedro.patron@seebyte.com}

 \author{Helen Hastie}
 \affiliation{%
  \institution{Heriot-Watt University, MACS}
   \city{Edinburgh} 
   \country{UK} 
   \postcode{EH14 4AS}
 }
\email{h.hastie@hw.ac.uk}




\renewcommand{\shortauthors}{Chiyah Garcia et al.}

\begin{abstract}
Autonomous systems in remote locations have a high degree of autonomy and there is a need to explain {\it what} they are doing and {\it why} in order to increase transparency and maintain trust. Here, we describe a natural language chat interface that enables vehicle behaviour to be queried by the user. We obtain an interpretable model of autonomy through  having an expert `speak out-loud' and provide explanations during a mission. This approach is agnostic to the type of autonomy model and as expert and operator are from the same user-group,
we predict that these explanations will align well with the  operator's mental model,
increase transparency  and assist with operator training. 
\end{abstract}

%
%



\maketitle

\section{Introduction}

Autonomous systems (AxV) now routinely operate in regions that are dangerous or impossible for humans to reach, such as the deep underwater environment. 
Typically, remote robots instil less trust than those co-located \cite{Bainbridge2008,hastieicmi2017trust}. This combined with high vulnerability in hazardous, high-stakes environments, such as that described in \cite{Hastie2018}, means that the interface between operator and AxV is key in maintaining situation awareness and understanding. Specifically, AxVs need to be able to maintain a continuous communication with regards to what they are doing; and increase transparency through explaining their actions and behaviours. 

Explanations can help formulate clear and accurate mental models of autonomous systems and robots. Mental models, in cognitive theory, provide one view on how humans reason either functionally (understanding what the robot does) or structurally (understanding how it works). Mental models are important as they strongly impact how and whether robots and systems are used. In previous work, explanations have been categorised as either explaining 1) machine learning as in \cite{Ribeiro2016} who showed that they can increase trust; 2) explaining plans \cite{Tintarev2014,Chakraborti2017}; 3) verbalising robot \cite{Rosenthal2016a} or agent rationalisation \cite{Ehsan2017}. However, humans do not present a constant verbalisation of their actions but they do need to be able to provide information on-demand about {\it what} they are doing and {\it why} during a live mission.





\begin{figure}[t]
\frame{\includegraphics[width=1.0\linewidth]{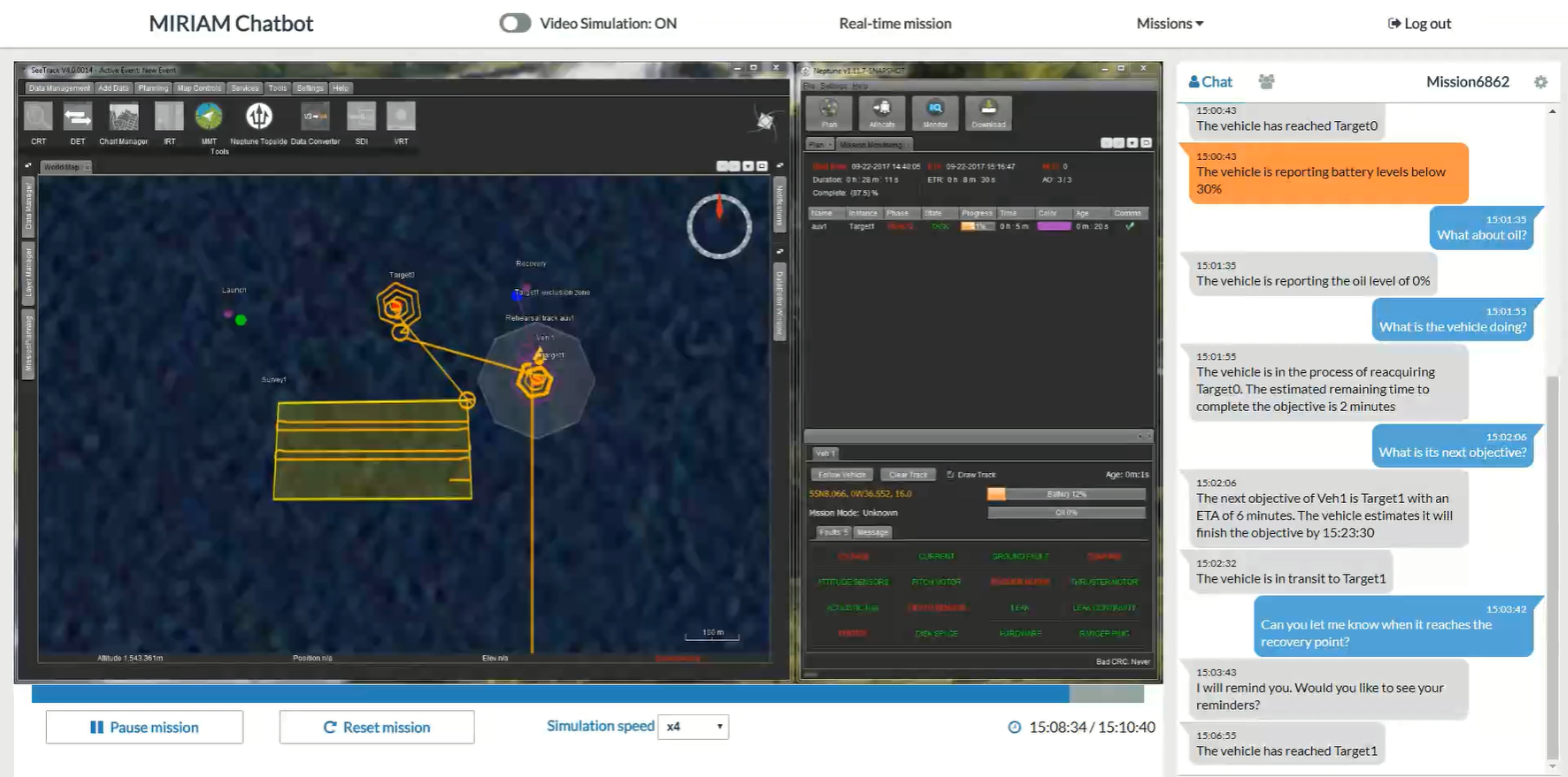}}
\caption{The MIRIAM multimodal interface with SeeTrack interface showing the predicted path of the vehicle on the left and the MIRIAM chat interface on the right}
\label{fig:miriam}
\end{figure}

We present here, MIRIAM, (Multimodal Intelligent inteRactIon for Autonomous systeMs), as seen in Figure  \ref{fig:miriam}. MIRIAM allows for these `on-demand' queries for status and explanations of behaviour. MIRIAM interfaces with the Neptune autonomy software provided by SeeByte Ltd and runs alongside their SeeTrack interface.  In this paper, we focus on explanations of behaviours and describe a method that is agnostic to the type of autonomy method. With respect to providing communication for monitoring, please refer to \cite{Hastie17demo} for further details and an overview of the system.


\section{Explanations for Remote Autonomy}
Types of explanations include {\it why} to provide a trace or reasoning and {\it why not} to elaborate on the system's control method or strategy \cite{GregorBenbasat}. Lim et al. (2009) \cite{Lim2009} show that both {\it why} and {\it why not} explanations increase understanding but only {\it why} increases trust. We adopt here the `speak-aloud' method whereby an expert provides rationalisation of the AxV behaviours while watching videos of missions on the SeeTrack software.  This has the advantage of being agnostic to the method of autonomy and could be used to describe rule-based autonomous behaviours but also complex deep models. Similar human-provided rationalisation has been used to generate explanations of deep neural models for game play \cite{Ehsan2017}.

\begin{figure}[t]
\frame{\includegraphics[width=1.0\linewidth]{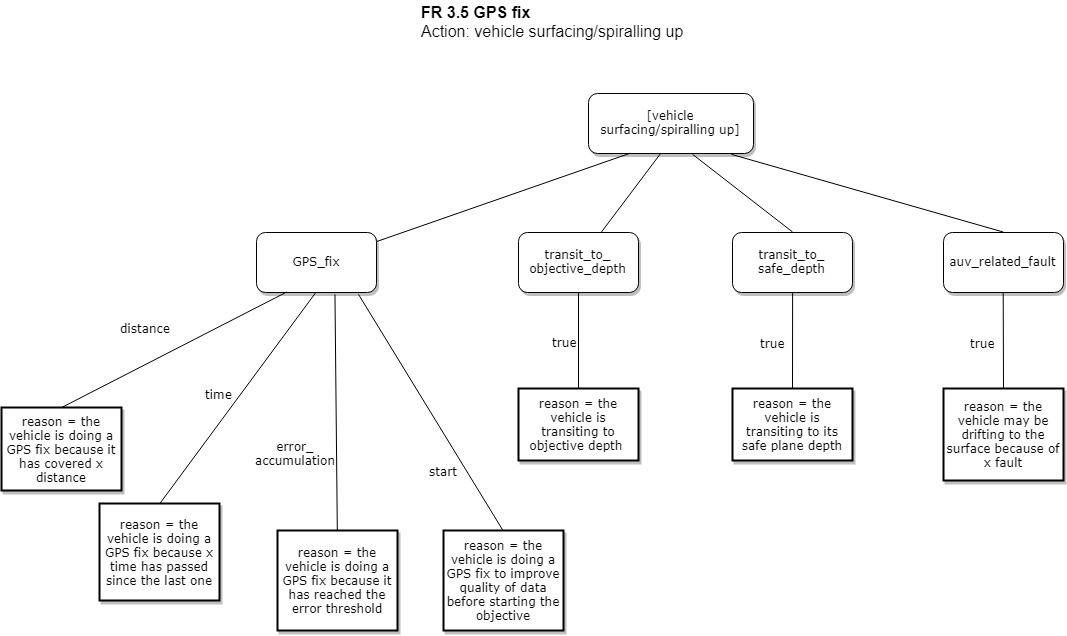}}
\caption{Autonomy model for vehicle surfacing}
\label{fig:tree}
\end{figure}

An interpretable model of autonomy was then derived from the expert, as partially shown in Figure \ref{fig:tree}. If a {\it why} request is made, the decision tree is checked against the current mission status and history and the possible reasons are determined, along with a probability. 
As we can see from example outputs in Figure \ref{fig:chatzoom}A, there may be multiple reasons with varying levels of certainty depending on the information available at a given point in the mission. Hence in this example, when the same {\it why} question is asked at a later point then only one higher confidence answer is given.

In the example scenario given in Figure \ref{fig:chatzoom}B, the operator is  able to observe in the SeeTrack interface that the vehicle has not done a GPS fix for some time. The user asks why it is not doing a GPS fix and the answer explains the relevant constraints on the vehicle, as captured in the interpretable autonomy model. The surface representations of the explanations are generated using template-based NLG. The wording of the output reflects the certainty on three levels: above 80\% (high), 80\%  to 40\% (medium) and below 40\% (low).

\section{Discussion and Future work}
\begin{figure}[ht]
\frame{\includegraphics[width=1\linewidth]{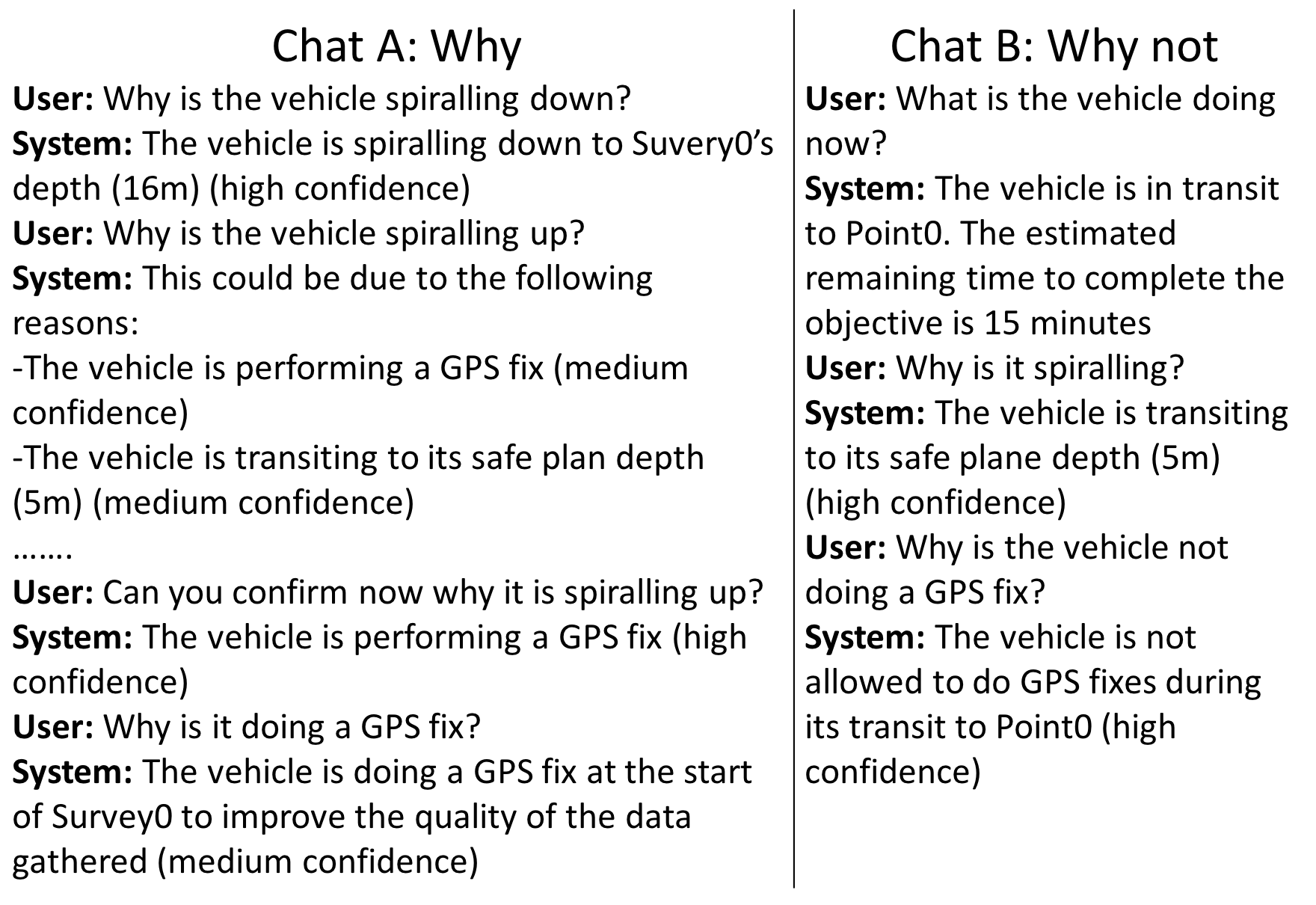}}
\caption{A: MIRIAM chat showing examples of {\it why} explanations; B showing examples {\it why not} explanations }
\label{fig:chatzoom}
\end{figure}

Future work includes conducting user evaluations to examine the trade-off between providing all of the information, even if one is not 100\% sure (`completeness') versus providing only those statements with very high confidence (`soundness').  The former is shown in Figure \ref{fig:chatzoom}. This trade-off will vary between personnel with different information needs and expertise  and domains as discussed in \cite{Kulesza2013}. 
Verbal indicators (e.g. "It is likely/probable") have been used in weather reporting to reflect levels of certainty \cite{wmo}. However, informal feedback from users indicate that the use of such verbal indicators may reduce confidence in the reporting system and therefore may not be suitable for highly critical, high risk situations. Exactly how these should be expressed is the subject of future work. 


We present here a method for explaining behaviours of remote AxV, which is agnostic to the autonomy model. Fortunately, in this domain, it is appropriate for the expert to be from the same pool of end-users (i.e. operators) and therefore explanations are likely to align with their mental models and assumptions about the system. This will not always be the case, as described in \cite{Chakraborti2017}, e.g. for in-home help robots, where users and planning experts have disparate mental models. Future work will involve the evaluation of explanations with respect to mental models. 


\begin{acks}
This research was funded by DSTL (REGIME 0913-C2-Ph2-009 and MIRIAM/ACC101939); EPSRC (EP/R026173/1, 2017-2021); RAEng/ Leverhulme Trust Senior Research Fellowship Scheme (Hastie/ LTSRF1617/13/37). Thanks to Prof. Y Petillot and Dr Z Wang.
\end{acks}

\bibliographystyle{ACM-Reference-Format}
\bibliography{bodyTHISONE-conf}

\end{document}